%% file: main.tex
\documentclass[a4paper]{article}

\usepackage{INTERSPEECH2019}
\usepackage{multirow}
\newcommand{\twolinecell}[2]{\vtop{\hbox{\strut #1}\hbox{\strut #2}}}

\title{Transformers with convolutional context for ASR}
\name{Abdelrahman Mohamed, Dmytro Okhonko, Luke Zettlemoyer}

\address{Facebook AI Research}
\email{{abdo,oxo,lsz}@fb.com}

\begin{document}

\maketitle
\begin{abstract}
The recent success of transformer networks for neural machine translation and other NLP tasks has led to a surge in research work trying to apply it for speech recognition. Recent efforts studied key research questions around ways of combining positional embedding with speech features, and stability of optimization for large scale learning of transformer networks. In this paper, we propose replacing the sinusoidal positional embedding for transformers with convolutionally learned input representations. These contextual representations provide subsequent transformer blocks with relative positional information needed for discovering long-range relationships between local concepts. The proposed system has favorable optimization characteristics where our reported results are produced with fixed learning rate of 1.0 and no warmup steps. The proposed model achieves a competitive 4.7\% and 12.9\% WER on the Librispeech ``test clean'' and ``test other'' subsets when no extra LM text is provided.\footnote{\texttt{Code available at: github.com/pytorch/fairseq/\newline tree/master/examples/speech\_recognition}}

\end{abstract}

\section{Introduction}

\input{intro.tex}

\section{Transformers with convolutional context}

\input{transformer.tex}

\section{Experimental results}

\input{experiments.tex}



\section{Conclusion and future work}

We presented a transformer seq2seq ASR system with learned convolutional context, both for the encoder and the decoder. Input convolutional layers capture relative positional information which enables subsequent transformer blocks to learn long range relationships between local concepts in the encoder, and recover the target sequence in the decoder. Using a deep transformer encoder was important to reach best performance, as we demonstrated empirically. Our best configuration achieves 12\% and 16\% relative reduction in WER compared to previously published systems on Librispeech ``dev other" and ``test other" subsets respectively, when no extra LM text is provided. Combining the proposed system with a better training procedure, e.g. Optimal Completion Distillation (OCD)\cite{ocd} is an interesting future avenue for combining observed WER gains.

\bibliographystyle{IEEEtran}
\bibliography{bib_short}

\end{document}

%% file: intro.tex
Speech Recognition systems have experienced many advances over the past decade, with neural acoustic and language models leading to impressive new levels of performance across many challenging tasks \cite{dnn_speech,tomas_2010}. Advances in alignment-free sequence-level loss functions like CTC and ASG \cite{ctc_graves, w2l_ronan} enabled easier training with letters as output units \cite{e2e_graves, deep_speech1}. The success of sequence-to-sequence models in neural machine translation systems \cite{cho_nmt, ioq_mt} offered further simplification to ASR systems by integrating the acoustic and the language models into a single encoder-decoder architecture that is jointly optimized \cite{las_16, attention_speech}. The encoder focuses on building acoustic representations that the decoder, through different attention mechanisms, can use to generate the target units. \\


Recently, transformer networks have been shown to perform well for neural machine translation \cite{transformer} and many other NLP tasks \cite{bert}. A Transformer layer distinguishes itself from a regular recurrent network by entirely relying on a key-value ``self"-attention mechanism for learning relationships between distant concepts, rather than relying on recurrent connections and memory cells to preserve information, as in LSTMs, that can fade over time steps. 
Transformer layers can be seen as bag-of-concept layers because they don't preserve location information in the weighted sum self-attention operation. To model word order, sinusoidal positional embeddings are used \cite{transformer}. 

There has been recent research interest in using transformer networks for end-to-end ASR both with CTC loss \cite{ctc_transformer} and in an encoder-decoder framework \cite{cmu_transformer, syllable_transformer} with modest performance compared to baseline systems. For a standard hybrid ASR system, \cite{Povey2018ATS} introduced a time-constrained key-value self-attention layer to be used in tandem with other TDNN and recurrent layers. Using time-restricted self-attention context enabled the authors to model input positions as 1-hot vectors, however, they didn't show a conclusive evidence for the impact of the self-attention context size. One interesting research question in all previous work was: how to best introduce positional information for input speech features. Answers range from dropping it altogether, adding it to input features/embedding, and concatenating it with input features leaving it to the neural network to decide how to combine them. \\


In this paper, we take an alternative approach. We propose replacing sinusoidal positional embedding with contextually augmented inputs learned by 2-D convolutional layers over input speech features in the encoder, and by 1-D convolutional layers over previously generated outputs in the decoder. Lower layers build atomic concepts, both in encoders and decoders, by learning local relationships between time steps. Long-range sequential structure modeling is left to subsequent layers. Although the transformer's flexible inductive bias is able to mimic convolution filters in its lower layers, we argue that this comes at the expense of brittle optimization. We believe that adding early convolutional layers allows the model to learn implicit relative positional encodings which enable subsequent transformer layers to recover the right order of the output sequence.\\

Using convolutional layers as input processors before recurrent layers in acoustic encoders has been previously proposed for computational reasons with minimal impact on performance \cite{espnet}. So, we focus our experiments on understanding the impact of the convolutional context size consumed by the decoder 1-D convolutional layers. 
Our best model configuration, with a fixed learning rate of 1.0, no hyperparameter or decoder optimization, achieves 12\% and 16\% relative reduction in WER compared to previously published results on the acoustically challenging Librispeech \cite{librispeech} ``dev other" and ``test other" subsets, when no extra LM text data is used during decoding.

%% file: transformer.tex
We propose dividing the modeling task into two sub-components: learning local relationships within a small context with convolutional layers, and learning global sequential structure of the input with transformer layers. This division simplifies transformer optimization leading to more stable training and better results because we don't need to force lower transformer layers to learn local dependencies. 

\subsection{Transformer layer}
Transformer layers \cite{transformer} have the ability to learn long range relationships for many sequential classification tasks \cite{bert}. Multi-head self-attention is the core component of transformer layers. Let $d_{input}$ be input dimension to a transformer layer, each time step in the input is projected into $d_k$, $d_k$, $d_v$ dimensional vectors representing the queries($Q$), keys($K$) and values($V$) for attention, where similarities between keys and queries determine combination weights of values for each time step,	
\begin{equation}
  Attention(Q, K, V) = Softmax(\dfrac{QK^T}{\sqrt{d_k}})V
\end{equation}
The dot product between keys and queries is scaled by the inverse square root of the key dimension. This self-attention operation is done $h$ times in parallel, for the case of $h$ attention heads, with different projection matrices from $d_{input}$ to $d_k$, $d_k$, and $d_v$.
 The final output is a concatenation of $h$ vectors each with dimension $d_v$ which is in turn linearly projected to the desired output dimension of the self-attention layer. 

\begin{figure}[t]
  \centering
  \includegraphics[width=\linewidth]{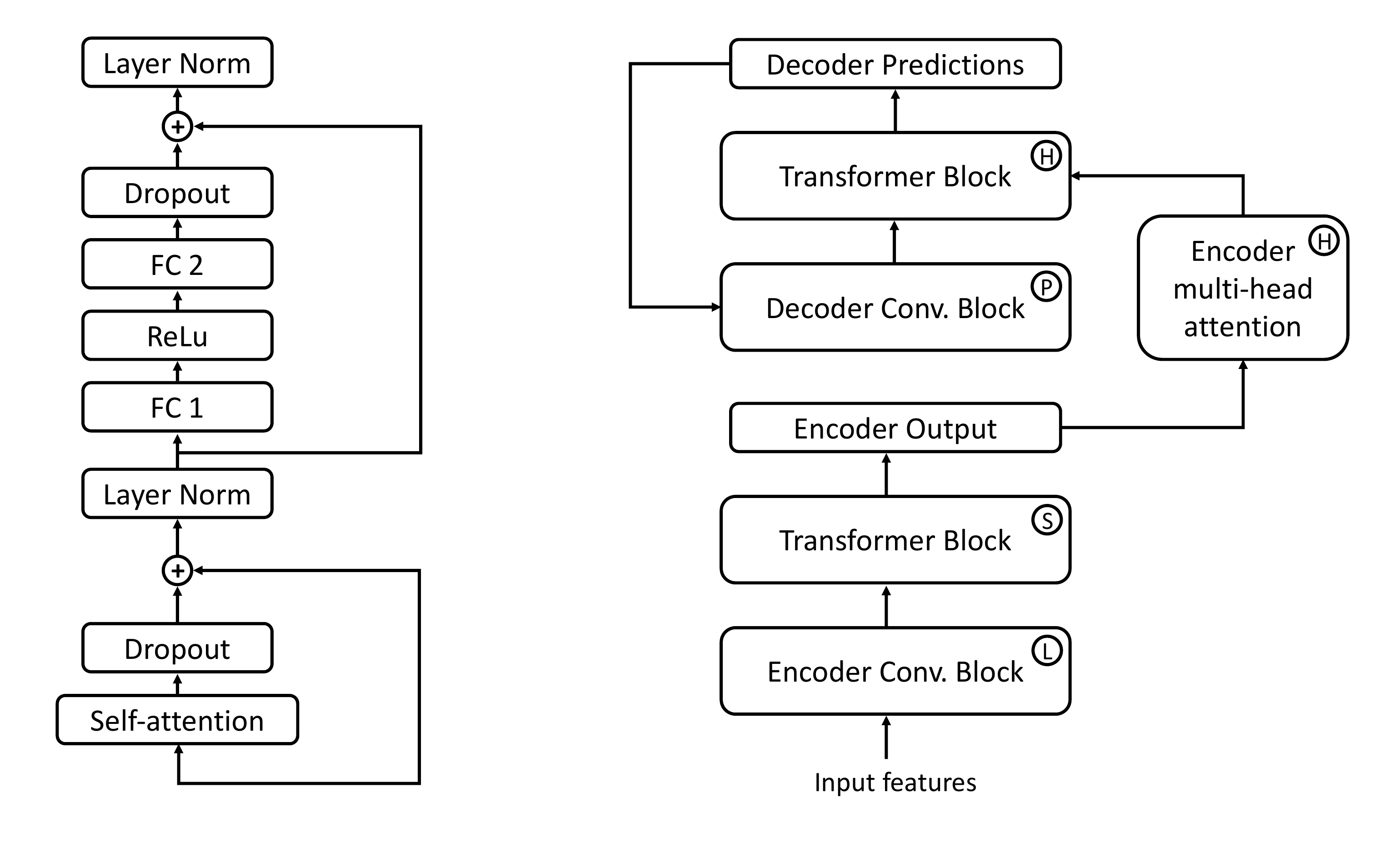}
  \caption{Left: components of one transformer block. Right: Block diagram of the full end-to-end model}
  \label{fig:transformer_full}
\end{figure}

On top of the self-attention component, transformer layers have multiple operations applied on each time step; dropout, residual connection, layer norm, two fully connected layers with a ReLU layer in between, another residual and Layer norm operations. Figure(\ref{fig:transformer_full})-left show the details of one transformer layer as proposed by \cite{transformer}.

\subsection{Adding context to transformer}

\begin{figure}[t]
  \centering
  \includegraphics[width=\linewidth]{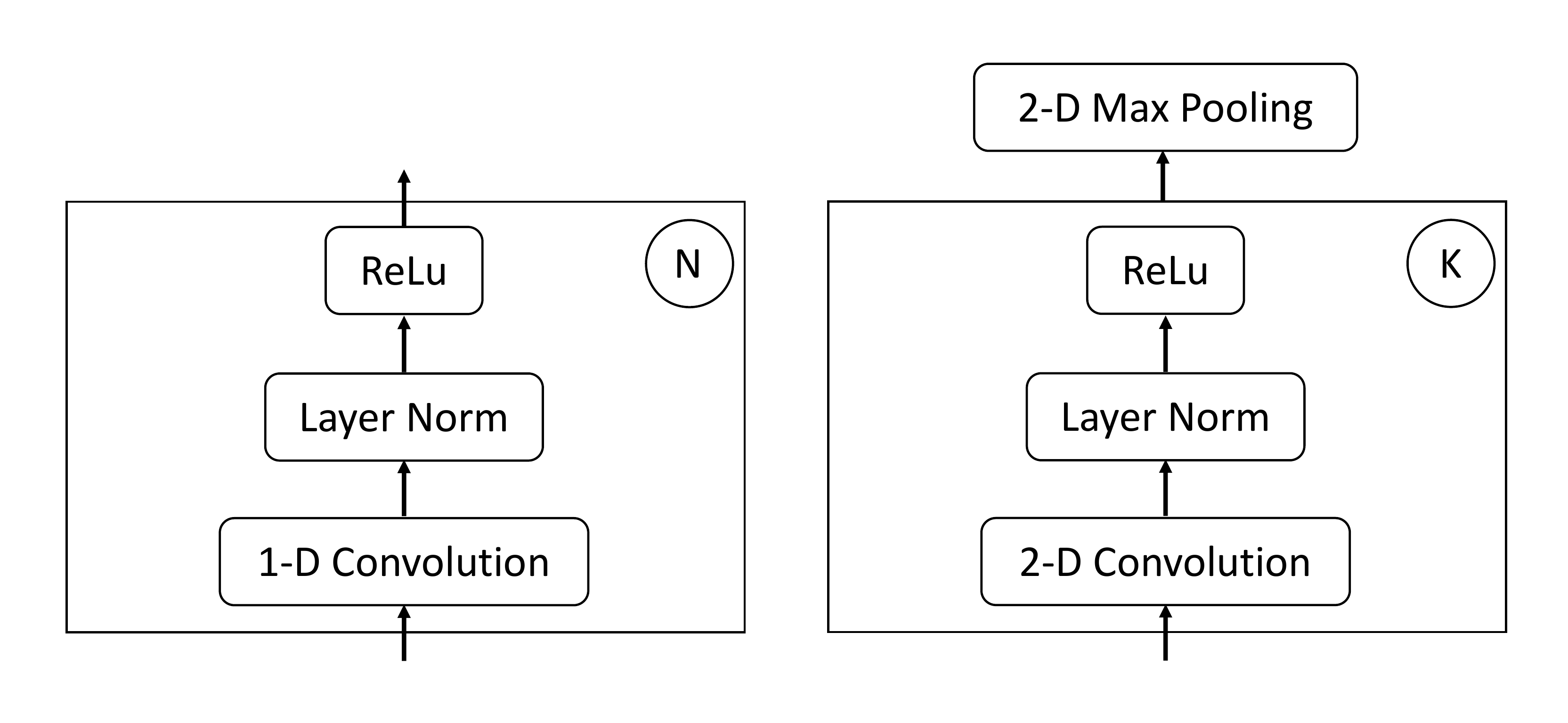}
  \caption{Left: One decoder side 1-D convolutional block. Right: One encoder side 2-D convolutional block.}
  \label{fig:conv_blocks}
\end{figure}

Our convolutional layers are added below the Transformer layers, and we do not make any use of positional encodings. 
The model learns an acoustic language model over the bag of discovered acoustic units as it goes deeper in the encoder. The experimental results show that using a relatively deep encoder is critical for getting good performance. 
For the encoder, we used 2-D convolutional blocks with layer norms and ReLU after each convolutional layer. Each convolutional block contains $K$ convolutional layers followed by a 2-D max pooling layer, as shown in figure(\ref{fig:conv_blocks})-right. For the decoder, we follow a similar approach using 1-D convolutions over embeddings of previously predicted words (shown in figure(\ref{fig:conv_blocks})-left with $N$ 1-D convolutional layers in each decoder convolutional block).

\subsection{Full end-to-end model architecture}
Figure(\ref{fig:transformer_full})-right shows the full end-to-end system architecture. 
Each block in the model is repeated multiple times (shown on the top right corner of each block). On the decoder side, we use a separate multi-head attention layer to aggregate encoder context for each decoder transformer block. We found that having more than one attention layer improves the overall system recognition performance. The decoder 1-D convolution only looks at historical predictions with its end point at the current time step. Similarly, the transformer layers have future target steps masked, so that decoder self-attention is only running over current and previous time steps to respect left-to-right output generation. We are not investigating online/streaming decoding conditions in this paper, so the encoder self-attention is allowed to operate over the entire input utterance.

%% file: experiments.tex
\subsection{Experimental Setup}

We evaluate performance on the Librispeech dataset \cite{librispeech} containing 1000h of training data with development and test sets split into simple (``clean") and harder (``other") subsets \footnote{We decided to concentrate on Librispeech and not on smaller datasets, e.g. TIMIT, WSJ, as with current model capacities research findings on smaller datasets can't reliably generalize to new scenarios and don't provide universal modeling trends. Early CTC experiments showed no gains for WSJ \cite{Graves13hybridspeech} while it was later proved to be one of the current best large scale loss functions \cite{deep_speech1}.}. We use 5k ``unigram" subword target units learned by the sentence piece package \cite{sent_piece} with full coverage of all training text data. Input speech is represented as 80-D log mel-filterbank coefficients plus three fundamental frequency features computed every 10ms with a 25ms window. 

All experiments were not tuned to best possible performance using training hyperparameter or decoder optimization. We don't use scheduled sampling or label smoothing. For regularization, we use a single dropout rate of 0.15 across all blocks as part of our default configurations. For model optimization, we use the AdaDelta algorithm \cite{adadelta} with fixed learning rate=1.0 and gradient clipping at 10.0. We run all configurations for 80 epochs, we then report results on an average model computed over the last 30 checkpoints. Averaging the last few checkpoints brings the model weights closer to the nearest local minimum. We could have stopped training models much earlier than 80 epochs, with a different early stopping point for different runs, but decided to stick by a generic training recipe to simplify reproducing our results. It is important to mention that we aren't using a learning rate warmup schedule and yet the model converges to the reported WER results in a stable way. This general fixed training recipe wasn't optimized on any part of Librispeech. 

The standard convolutional tranformer model used in most experiments has the following configuration: (1) Two 2-D convolutional blocks, each with two conv. layers with kernel size=3, max-pooling kernel=2. The first block has 64 feature maps while the second has 128, (2) 10 encoder transformer blocks all with transformer dim=1024, 16 heads, intermediate ReLU layer size=2048, (3) decoder input word embedding dim=512, (4) three 1-D conv. layers each with kernel size=3, no max pooling is used for the decoder side 1-D convolution, and (5) 10 decoder transformer blocks each with encoder-side multihead attention, otherwise the configuration is identical to the encoder transformer block. This canonical model has about 223M parameters, and it takes about 24 hours to perform all 80 epochs on 2 machines each with 8GPUs with 16GB of memory. All results are reported without any external language model trained on extra text data. Our focus is to study the contextual transformer decoder's ability to model the statistical properties of the spoken training data. We use beam size of 5 during inference for all experiments except mentioned otherwise.

\begin{table}[h!]
\centering
\begin{tabular}{ c||c|c|c|c } 
 Model &  \twolinecell{dev}{clean} & \twolinecell{dev}{other} & \twolinecell{test}{clean} & \twolinecell{test}{other} \\
 \hline \hline
Conv. context & 5.2& 13.7& 5.3& 14.0 \\
Sin. pos. embd & 5.8 & 14.2 & 5.4 & 14.8 \\
(1) + (2) & 5.2& 13.8& 5.3& 14.0 \\
\hline
One layer of enc. att. & 6.4 & 15.2 & 6.3 & 15.9\\
\hline
32 heads in enc/dec&5.3&14.1&5.4&14.6\\
\hline
4k transformer ReLU&5.3&13.2&5.3&13.4\\
  \hline \hline
\end{tabular}
\caption{The impact of different modeling decision on WER}
 \label{tab:compare_to_pos}
\end{table}

\subsection{Model Comparisions}

We first studied performance of our approach to alternative architectures and positional encoding schemes. In table(\ref{tab:compare_to_pos}) we show the WER of the proposed transformer encoder-decoder model with convolutional context using the canonical configuration in the first row. Replacing the 1-D convolutional context in the decoder with sinusoidal positional embedding, as proposed in the baseline machine translation transformers \cite{transformer} and adopted in \cite{ctc_transformer, syllable_transformer}, shows inferior WER performance. By combining sinusoidal and convolutional position embedding (rows 1+2), we don't observe any gains. This supports our intuition that the relative convolutional positional information provides sufficient signal for the transformer layers to recreate more global word order. We also found that having multiple encoder-side attention layers is critical for achieving the best WER. Increasing the intermediate ReLU layer in each encoder and decoder layer was found to  greatly improve the overall WER across different sets, however increasing the number of attention heads, while keeping the attention dimension the same, deteriorates the performance.

To better understand these results, we also studied the effects of different hyperparameter settings.
Table(\ref{tab:DecConv}) shows the effect of different decoder convolutional context sizes spread over different depths. All configurations in table(\ref{tab:DecConv}) share the same canonical configuration and the number of 1-D conv. feature maps were chosen to ensure the total number of parameters are fixed between all configurations. The best performance comes from using the same parameter budget over wider context that is built over multiple convolutional layers. However, the decoder is able to get reasonable WER even with a context of just 3 words as input to the transformer layers.
\begin{table}[h!]
\centering
\begin{tabular}{ c||c||c|c|c|c } 
\twolinecell{Conv.}{depth} & \twolinecell{Cxt size}{(num. kernels)} & \twolinecell{dev}{clean} & \twolinecell{dev}{other} & \twolinecell{test}{clean} & \twolinecell{test}{other} \\
 \hline \hline
 \multirow{5}{3em}{1} &3 (3)  & 5.3 & 13.8 & 5.4 & 14.1 \\
&5 (5)  &5.4 & 14.1 & 5.5 & 14.0 \\
&7 (7)  & 5.4 & 13.9 & 5.4 & 14.5 \\
&9 (9)  & 5.4 & 13.9 & 5.5 & 14.0 \\
&11 (11) & 5.3 & 13.6 & 5.4 & 13.8 \\
\hline \hline  
 \multirow{4}{3em}{2} &5 (3-3)& 5.3 & 14.0 & 5.2 & 14.5 \\
&7 (3-5)&  5.5 & 14.7 & 5.9 & 14.8 \\
&9 (5-5)& 5.2 & 13.9 & 5.6 & 14.2 \\
&11 (5-7)&  5.2& 14.1 & 5.4 & 14.6 \\
\hline \hline  
 \multirow{3}{3em}{3} 
&7 (3-3-3)& 5.2& 13.7& 5.3& 14.0 \\
&9 (3-3-5)& 5.3& 13.8& 5.4& 14.1 \\
&11 (3-5-5) & 5.6 &14.3 &5.4&14.2 \\
\hline \hline  
 \multirow{2}{3em}{4} 
&9 (3-3-3-3)& 5.0 & 13.5 & 5.4 & 13.9 \\
&11 (3-3-3-5)&  5.0 & 13.6 & 5.2 & 13.7 \\
\hline \hline
\end{tabular}
\caption{WER for different decoder convolution architectures}
 \label{tab:DecConv}
\end{table}
Using a deep transformer encoder capture long range structure of the data as an acoustic LM built on top of learned concepts from the convolutional layers. Also deeper encoder help marginalize out global utterance specific speaker and environment characteristics while focusing on the content. A deeper deocder, although not as critical, showed better overall performance. Table(\ref{tab:depth}) shows WER for different depth configurations. We wanted to understand the effect of fixing encoder depth while changing decoder and vice versa, fixing the total sum of encoder and decoder depths, as well as using same depth on both sides all the way up to 14 transformer layers.
\begin{table}[h!]
\centering
\begin{tabular}{ c||c||c|c|c|c } 
 Setup & \twolinecell{Enc/Dec}{depth} & \twolinecell{dev}{clean} & \twolinecell{dev}{other} & \twolinecell{test}{clean} & \twolinecell{test}{other} \\
 \hline \hline
 \multirow{5}{3em}{Same Enc/Dec depth} & 6/6&   5.6 & 14.5 & 5.7 & 15.3 \\
 &8/8&   5.3& 14.0& 5.3& 13.9\\
 &10/10& 5.2& 13.7& 5.3& 14.0\\
 &12/12& 5.0& 13.0& 5.0& 13.3\\
 &14/14& 5.0& 12.9& 5.0& 13.4\\
\hline \hline  
 \multirow{5}{3em}{Same total depth} &2/10& 7.5& 18.4& 7.8& 18.7 \\
&4/8&  6.2& 15.5& 6.0& 16.0 \\
&6/6&  5.6& 14.5& 5.7& 15.3 \\
&8/4&  5.4& 13.9& 5.3& 14.3 \\
&10/2& 5.2& 14.1& 5.4& 14.6 \\
\hline \hline  
 \multirow{5}{3em}{Same Enc depth} &10/2&  5.2& 14.1& 5.4& 14.6 \\
&10/4&  5.1& 13.7& 5.2& 14.2 \\
&10/6&  5.0& 13.3& 5.1& 13.7 \\
&10/8&  5.2& 13.6& 5.2& 14.3 \\
&10/10& 5.2& 13.7& 5.3& 14.0 \\
\hline \hline  
 \multirow{5}{3em}{Same Dec depth} &2/10&  7.5& 18.4& 7.8& 18.7 \\
&4/10&  6.4& 15.4& 6.3& 16.4 \\
&6/10& 5.7& 14.5& 5.7& 14.6 \\
&8/10& 5.2& 14.0& 5.4& 14.3 \\
&10/10& 5.2& 13.7& 5.3& 14.0 \\
\hline  \hline
\end{tabular}
\caption{WER for different transformer depth in encoder and decoder}
 \label{tab:depth}
\end{table}

\subsection{Final Results}
Based on these experimental findings, we combined the best performing configurations into one model that is similar to the canonical model except: (1) we use 4k ReLU layers in all transformer blocks in the encoder and the decoder,  (2) we use 16 encoder transformer blocks, and (3) we only use 6 decoder transformer blocks. The results of the best model are shown in table(\ref{tab:prev_sys}). For decoding of the best model we used beam size of 20.

Table(\ref{tab:prev_sys}) compares this model to other previously published results on the Librispeech dataset. For completeness, we added models that use externally trained LMs on extra text data, although their results aren't comparable to ours. Compared to models with no external LM, our model brings 12\% to 16\% relative WER reduction on the acoustically challenging ``dev other" and ``test other" subsets of Librispeech. This suggests that the convolutional tranformer indeed learns long-range acoustic characteristics of speech data, e.g. speaker and environment characteristics, because the model doesn't bring much improvement to the ``dev clean" and ``test clean" subsets which need external text data for improvement. The results confirm our belief that the improvements found in this paper are orthogonal to further potential improvements to the WER using an LM trained on much larger text corpus. 

\begin{table}[h!]
\centering
\setlength\tabcolsep{4.0pt}
\begin{tabular}{ c||c||c|c|c|c } 
 Model & \twolinecell{LM on}{extra text} & \twolinecell{dev}{clean} & \twolinecell{dev}{other} & \twolinecell{test}{clean} & \twolinecell{test}{other} \\
 \hline \hline
\twolinecell{CAPIO}{spk adpt\cite{CAPIO}}&RNNLM&3.12&8.28&3.51&8.58\\
\hline
LSTM\cite{achen_s2s}&4gramLM&4.79&14.31&4.82&15.30\\
Gated Cnv\cite{vitaliy_2017}&4gramLM&4.6&13.8&4.8&14.5\\
\twolinecell{Tsf w/sin}{pos embd\cite{ctc_transformer}}&4gramLM&-&-&4.8&13.1\\
TDS Cnv\cite{Hannun19}&4gramLM&3.75&10.70&4.21&11.87\\
\hline
LSTM\cite{achen_s2s}&LSTMLM&3.54&11.52&3.82&12.76\\
\hline
Fully Cnv\cite{full_conv_asr}&ConvLM&3.16&10.05&3.44&11.24\\
TDS Cnv\cite{Hannun19}&ConvLM&3.01&8.86&3.28&9.84\\
\hline \hline
TDS Cnv\cite{Hannun19}&None&5.04&14.45&5.36&15.64\\
LSTM\cite{achen_s2s}&None&4.87&14.37&4.87&15.39\\
\hline
LSTM (MLE)\cite{ocd}&None&-&-&5.7&15.4\\
LSTM (OCD)\cite{ocd}&None&-&-&\textbf{4.5}&13.3\\
\hline
\twolinecell{Cnv Cxt Tsf (MLE)}{\textbf{(ours)}}&None&\textbf{4.8}&\textbf{12.7}&4.7&\textbf{12.9}\\
\hline \hline
\end{tabular}
\caption{WER comparison with other previously published work on Librispeech}
 \label{tab:prev_sys}
\end{table}